\documentclass[]{article}
\usepackage{lmodern}
\usepackage{amssymb,amsmath}
\usepackage{ifxetex,ifluatex}
\usepackage{fixltx2e} 
\ifnum 0\ifxetex 1\fi\ifluatex 1\fi=0 
  \usepackage[T1]{fontenc}
  \usepackage[utf8]{inputenc}
\else 
  \ifxetex
    \usepackage{mathspec}
  \else
    \usepackage{fontspec}
  \fi
  \defaultfontfeatures{Ligatures=TeX,Scale=MatchLowercase}
\fi
\IfFileExists{upquote.sty}{\usepackage{upquote}}{}
\IfFileExists{microtype.sty}{%
\usepackage[]{microtype}
\UseMicrotypeSet[protrusion]{basicmath} 
}{}
\PassOptionsToPackage{hyphens}{url} 
\usepackage[unicode=true]{hyperref}
\PassOptionsToPackage{usenames,dvipsnames}{color} 
\hypersetup{
            pdftitle={Disparity Image Segmentation For ADAS},
            pdfauthor={Viktor Mukha \& Inon Sharony, L4B Automotive},
            colorlinks=true,
            linkcolor=Maroon,
            citecolor=Blue,
            urlcolor=blue,
            breaklinks=true}
\usepackage[margin=1in]{geometry}
\usepackage{longtable,booktabs}
\IfFileExists{footnote.sty}{\usepackage{footnote}\makesavenoteenv{long table}}{}
\usepackage{graphicx,grffile}
\makeatletter
\def\maxwidth{\ifdim\Gin@nat@width>\linewidth\linewidth\else\Gin@nat@width\fi}
\def\maxheight{\ifdim\Gin@nat@height>\textheight\textheight\else\Gin@nat@height\fi}
\makeatother
\setkeys{Gin}{width=\maxwidth,height=\maxheight,keepaspectratio}
\IfFileExists{parskip.sty}{%
\usepackage{parskip}
}{
\setlength{\parindent}{0pt}
\setlength{\parskip}{6pt plus 2pt minus 1pt}
}
\providecommand{\tightlist}{%
  \setlength{\itemsep}{0pt}\setlength{\parskip}{0pt}}
\ifx\paragraph\undefined\else
\let\oldparagraph\paragraph
\renewcommand{\paragraph}[1]{\oldparagraph{#1}\mbox{}}
\fi
\ifx\subparagraph\undefined\else
\let\oldsubparagraph\subparagraph
\renewcommand{\subparagraph}[1]{\oldsubparagraph{#1}\mbox{}}
\fi

\makeatletter
\def\fps@figure{htbp}
\makeatother

\title{Disparity Image Segmentation For ADAS}
\author{Viktor Mukha \&
\href{mailto:Inon.S@L4B-Software.com?subject=disparity_image_segmentation_for_adas}{Inon
Sharony}, L4B Automotive}
\date{2018-06-05}

\begin{document}
\maketitle

{
\hypersetup{linkcolor=black}
\setcounter{tocdepth}{3}
\tableofcontents
}
\section{Abstract}\label{abstract}

We present a simple solution for segmenting grayscale images using
existing Connected Component Labeling (CCL) algorithms (which are
generally applied to binary images), which was efficient enough to be
implemented in a constrained (embedded automotive) architecture. Our
solution customizes the region growing and merging approach, and is
primarily targeted for stereoscopic disparity images where nearer
objects carry more relevance. We provide results from a standard OpenCV
implementation for some basic cases and an image from the
\href{http://www.cvlab.cs.tsukuba.ac.jp/dataset/tsukubastereo.php}{Tsukuba
stereo-pair dataset}.

\section{Introduction}\label{introduction}

In various computer vision contexts, we are required to perform Blob
Extraction or Image Segmentation on an image. This image may be binary
(i.e.~each pixel either black or white), or not. ADAS usually entails
some sort of sensor fusion, with two common sensors being thermal
imaging and stereoscopic cameras. Both these sensors output false
``color'' grayscale images, in which pixel value expresses either the
temperature or parallax (also known as disparity), respectively.

In the following we will give examples from Image Segmentation of a
Disparity Image. The higher the disparity value, the closer the object.
An object at an infinite distance away from the camera will have a
disparity value of zero. In the context of Advanced Driver Assistance
Systems (ADAS), we use disparity image segmentation to detect objects in
the vehicle's path, primarily of interest are the closer objects (this
consideration will guide us throughout the following). In thermal
imaging we are also interested only in foreground objects, but can bring
more domain knowledge to bear in thresholding a certain part of the
temperature spectrum (in contrast, we don't know the relevant disparity
values in advance).

\subsection{Disparity image
segmentation}\label{disparity-image-segmentation}

The task of image segmentation in our context is to extract the blobs
from a post-processed disparity image. Each blob is described using a
bounding box, a rectangle tightly encompassing the blob.

Additionally, this module determines the maximum disparity value in the
bounding box and assigns it to the object description \texttt{struct}.

It is hard to reliably estimate the ``thickness'' of an object from
stereo-derived depth maps, so we should rather concentrate on the most
important distance for us: the closest distance to the object. If the
maximum value is found too rough an estimate, we should throw away
outliers by doing something similar to median filtering, but only in
order to find the value of interest. That is, we don't need to generate
a full matrix of disparity values for each rectangle. Basically, this is
a direct estimation of the distance to the object, so this will be
thoroughly investigated elsewhere.

The algorithm sorts the results in the order of decreasing disparity
value, since we are mostly interested in the closer objects.

\subsection{Previous work}\label{previous-work}

CCL is well researched Image Segmentation method, with many efficient
implementations. However, the vast majority of Image Segmentation
algorithms such as
\href{https://www.sciencedirect.com/science/article/pii/S0031320317301693}{Connected
Component Labeling (CCL)} only work on a binary image, while a disparity
image is a grayscale image.

There is one work which has shown that it is possible to extend
different CCL algorithms to grayscale images, by
\href{http://paper.ijcsns.org/07_book/200806/20080605.pdf}{Yapa \&
Harada}. However the implementation is not publicly available, and these
ideas have been implemented in \texttt{MATLAB} and are lacking proper
evaluation for embedded systems. Therefore, we have decided to derive a
more portable solution.

\subsection{Our attempts}\label{our-attempts}

The first try was to use the
\href{https://docs.opencv.org/3.3.1/d0/d7a/classcv_1_1SimpleBlobDetector.html\#details}{\texttt{cv::SimpleBlobDetector}}
class, which is the only blob extraction algorithm in OpenCV which works
on grayscale images instead of binary images. It uses multiple
thresholds for generating binary images and then applies the
\href{https://docs.opencv.org/2.4/modules/imgproc/doc/structural_analysis_and_shape_descriptors.html?highlight=findcontours\#findcontours}{\texttt{cv::findContours()}}
algorithm to binary images.
\href{https://docs.opencv.org/2.4/modules/imgproc/doc/structural_analysis_and_shape_descriptors.html?highlight=findcontours\#findcontours}{\texttt{cv::findContours()}},
in turn, uses a border following algorithm by
\href{http://download.xuebalib.com/xuebalib.com.17233.pdf}{Suzuki \&
Abe}.

The results of directly applying
\href{https://docs.opencv.org/3.3.1/d0/d7a/classcv_1_1SimpleBlobDetector.html\#details}{\texttt{cv::SimpleBlobDetector}}
to the industry-standard
\href{http://www.cvlab.cs.tsukuba.ac.jp/dataset/tsukubastereo.php}{Tsukuba
ground truth disparity images} were not satisfactory. The following are
results for parameters which allow all found keypoints to be shown. On
the left is the original disparity image, and on the right is a negated
image (since by default the algorithm looks for dark circles in bright
image):

\begin{longtable}[]{@{}ll@{}}
\toprule
Original & Negated\tabularnewline
\midrule
\endhead
\includegraphics{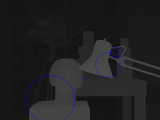} &
\includegraphics{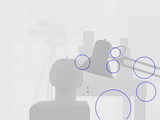}\tabularnewline
\bottomrule
\end{longtable}

Tweaking all parameters only filtered some of these keypoints away and
did not help finding other keypoints.

Thus, we decided to implement our own multiple-threshold algorithm using
OpenCV's existing CCL implementations.

\section{Solution}\label{solution}

The idea behind the algorithm is fairly simple: We apply a binarizing
threshold, run CCL, extract objects, and repeat this with a lower
threshold, until objects start to merge or we do not detect any objects
anymore. We decided not to use the
\href{https://ieeexplore.ieee.org/document/4310076}{histogram of the
grayscale image} since we are interested in the the nearest objects more
than the shape of the global histogram.

\subsection{Algorithm: Adaptive Non-binary
CCL}\label{algorithm-adaptive-non-binary-ccl}

\begin{enumerate}
\def\labelenumi{\arabic{enumi}.}
\tightlist
\item
  Determine the range of disparity in the input disparity image using
  \texttt{cv::minMaxLoc} and take \texttt{thresholdStepSize} (percent)
  of it as a threshold step.
\item
  Initialize an empty \texttt{storedObjects} vector.
\item
  Apply a binary threshold using \texttt{cv::threshold} to get a binary
  image for CCL.
\item
  Run CCL with this binary image.
\item
  Update the vector of stored objects with the newly detected objects
  using the following criteria (in this order):

  \begin{itemize}
  \tightlist
  \item
    if the new object is almost fully inside the bounding box of any
    stored object, discard it;
  \item
    if the new object does not intersect any of the stored objects, add
    it;
  \item
    if the new object contains the centers of more than one stored
    objects, this means two stored objects have merged, so discard this
    new object;
  \item
    if an old object is almost fully inside of any new object and the
    center of stored object is inside it too, update the stored object's
    bounding box and center with those from the new object.
  \end{itemize}
\end{enumerate}

\subsection{Implementation}\label{implementation}

The algorithm is implemented as a single pass over the threshold values,
then a single pass over any new objects each of which is running over
all stored objects to check the criteria. When checking for a merge,
complexity is added since we need to check if any other stored object's
center is inside of the new object. This adds another pass over the
stored objects.

The main factors of complexity are therefore the number of objects we
want to detect and the stopping criteria.

The ``blobs'' output is written to
\texttt{std::vector\textless{}DetectedObjects\textless{}int\textgreater{}\textgreater{}}.
Basically, a \texttt{DetectedObject} contains the 2D coordinates of the
center, the disparity value, and a bounding box (\texttt{cv::Rect}).
Please note that
\texttt{cv::Rect\_\textless{}unsigned\ short\textgreater{}} is
ill-defined for images with a dimension larger than \texttt{ushort}
pixels, since the intersection operator \texttt{\&} can overflow the
buffer. This is the reason for choosing the \texttt{int} data type.

The implemented algorithm is iterative in nature because it considers
objects from each of the previous steps. Under some circumstances (zero
value of the \texttt{numSameIterationsToStop} parameter, in the
following) we can parallelize the CCL computation using, e.g.
\texttt{cv::parallel\_for()}.

For the rest of the algorithm logic, we would still need to process all
objects detected by CCL sequentially, so before parallelizing, it is
important to profile how much time running CCL takes, and how much time
the logic takes. If both are taking too much time, it might be better to
opt for another approach from
\href{http://paper.ijcsns.org/07_book/200806/20080605.pdf}{Yapa \&
Harada}.

\subsection{Parameters}\label{parameters}

\begin{longtable}[]{@{}lllll@{}}
\toprule
Position & Name & Type & Domain & Default\tabularnewline
\midrule
\endhead
1 & \texttt{thresholdStepSize} & \texttt{float} & \([0,1]\) &
0.05\tabularnewline
2 & \texttt{numSameIterationsToStop} & \texttt{uchar} & \([0,2^8)\) &
0\tabularnewline
3 & \texttt{minObjDimension} & \texttt{ushort} &
\([0, \max(\textrm{width},\textrm{height}) - 1\) & 10\tabularnewline
4 & \texttt{maxObjDimension} & \texttt{ushort} &
\([0, \max(\textrm{width},\textrm{height}) - 1\) & 400\tabularnewline
5 & \texttt{commonAreaToConsiderBackground} & \texttt{float} & \([0,1]\)
& 0.9\tabularnewline
6 & \texttt{commonAreaToConsiderGrowing} & \texttt{float} & \([0,1]\) &
0.9\tabularnewline
\bottomrule
\end{longtable}

The rest of the parameters are the arguments to
\href{https://docs.opencv.org/3.2.0/d3/dc0/group__imgproc__shape.html\#ga107a78bf7cd25dec05fb4dfc5c9e765f}{\texttt{cv::connectedComponentsWithStats()}}
CCL method:

\begin{longtable}[]{@{}lllll@{}}
\toprule
Position & Name & Type & Domain & Default\tabularnewline
\midrule
\endhead
7 & \texttt{connectivity} & \texttt{int} & 8 or 4 & 8\tabularnewline
8 & \texttt{ltype} & \texttt{int} & \texttt{CV\_32SC1} or
\texttt{CV\_16UC1} & \texttt{CV\_16UC1}\tabularnewline
9 & \texttt{ccltype} & \texttt{int} & see the
\href{https://docs.opencv.org/3.2.0/d3/dc0/group__imgproc__shape.html\#ga5ed7784614678adccb699c70fb841075}{\texttt{cv::ConnectedComponentsAlgorithmsTypes}}
& \texttt{CCL\_GRANA}\tabularnewline
\bottomrule
\end{longtable}

\subsubsection{Definitions}\label{definitions}

\begin{enumerate}
\def\labelenumi{\arabic{enumi}.}
\tightlist
\item
  \texttt{thresholdStepSize}: The algorithm proceeds in descending order
  of disparity values moving the threshold by this step size. The
  default is 5\% of the disparity range (maximum disparity minus minimum
  disparity). Smaller steps help to separate objects which are close in
  depth to each other, bigger steps save calculation time since this
  parameter directly determines the number of CCL runs.
\item
  \texttt{numSameIterationsToStop}: This is a stop criteria of the
  algorithm. The default value of \texttt{0} lets the algorithm go over
  the complete range of disparity values, which is slow, and will only
  help to detect objects with lower disparity, which are probably not
  that important in our context. A positive-definite value will stop if
  nothing changed in the last \texttt{numSameIterationsToStop} steps.
  This parameter is to be adjusted together with
  \texttt{thresholdStepSize} if the algorithm takes too long to go over
  all thresholds.
\item
  \texttt{minObjDimension}: Minimal object dimension in pixels. Both
  height and width of the detected blob must satisfy this criteria for
  an object to be stored in the list of detected objects. Note that very
  small dimensions will allow the algorithm to accept noise as objects.
\item
  \texttt{maxObjDimension}: Maximal object dimension in pixels. Both
  height and width of a detected blob must satisfy this criteria for an
  object to be stored in the list of detected objects. Note that setting
  this parameter too low will make the detection of objects which are
  close to the cameras impossible. The default value is one less than
  the minimal frame dimension (height or width), so that the complete
  frame will not be recognized as an object.
\item
  \texttt{commonAreaToConsiderBackground}: Criteria for discarding a
  \emph{newly detected} object if it is at least
  \texttt{commonAreaToConsiderBackground} (ratio) inside of any
  \emph{stored} object. In most cases, such a detected object is behind
  an already detected object. This parameter is only needed for
  compensating for slightly shifted object detection.
\item
  \texttt{commonAreaToConsiderGrowing}: Criteria for updating a
  \emph{stored} object if it is at least
  \texttt{commonAreaToConsiderGrowing} (ratio) inside of a \emph{newly
  detected} object. This means that the object is growing. This
  parameter is only needed for compensating for slightly shifted object
  detection.
\item
  \texttt{connectivity}: 8-way or 4-way connectivity, respectively.
\item
  \texttt{ltype}: Output image label type. Currently \texttt{CV\_32S}
  and \texttt{CV\_16U} are supported.
\item
  \texttt{ccltype}: Connected components algorithm type.
\end{enumerate}

\section{Results}\label{results}

We have implemented four unit test cases for segmentation. The following
are the resulting images with rectangles around detected objects in
corresponding order:

\begin{longtable}[]{@{}llll@{}}
\toprule
\begin{minipage}[b]{0.04\columnwidth}\raggedright\strut
Test case\strut
\end{minipage} & \begin{minipage}[b]{0.09\columnwidth}\raggedright\strut
Description\strut
\end{minipage} & \begin{minipage}[b]{0.29\columnwidth}\raggedright\strut
Result\strut
\end{minipage} & \begin{minipage}[b]{0.47\columnwidth}\raggedright\strut
Evaluation\strut
\end{minipage}\tabularnewline
\midrule
\endhead
\begin{minipage}[t]{0.04\columnwidth}\raggedright\strut
1\strut
\end{minipage} & \begin{minipage}[t]{0.09\columnwidth}\raggedright\strut
3 filled rectangles of different disparity values, and 1 black rectangle
(faking failed disparity)\strut
\end{minipage} & \begin{minipage}[t]{0.29\columnwidth}\raggedright\strut
\includegraphics{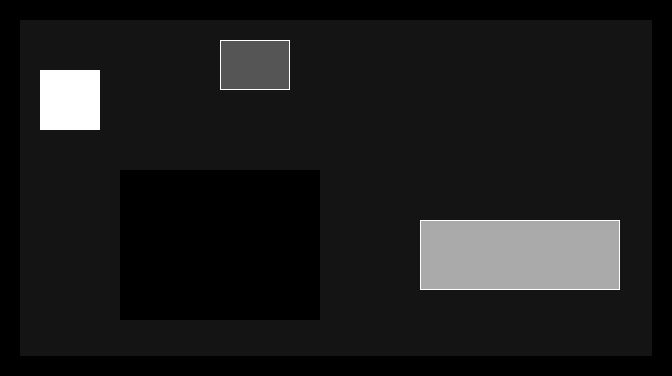}\strut
\end{minipage} & \begin{minipage}[t]{0.47\columnwidth}\raggedright\strut
3 objects are recognized correctly\strut
\end{minipage}\tabularnewline
\begin{minipage}[t]{0.04\columnwidth}\raggedright\strut
2\strut
\end{minipage} & \begin{minipage}[t]{0.09\columnwidth}\raggedright\strut
3 filled circles\strut
\end{minipage} & \begin{minipage}[t]{0.29\columnwidth}\raggedright\strut
\includegraphics{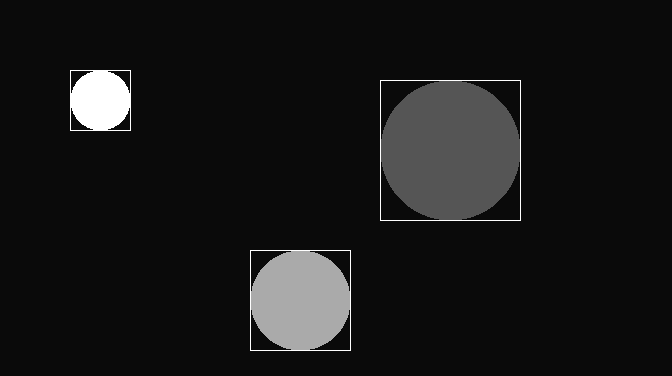}\strut
\end{minipage} & \begin{minipage}[t]{0.47\columnwidth}\raggedright\strut
3 objects are recognized correctly\strut
\end{minipage}\tabularnewline
\begin{minipage}[t]{0.04\columnwidth}\raggedright\strut
3\strut
\end{minipage} & \begin{minipage}[t]{0.09\columnwidth}\raggedright\strut
Disparity Ground Truth image from Tsukuba dataset\strut
\end{minipage} & \begin{minipage}[t]{0.29\columnwidth}\raggedright\strut
\includegraphics{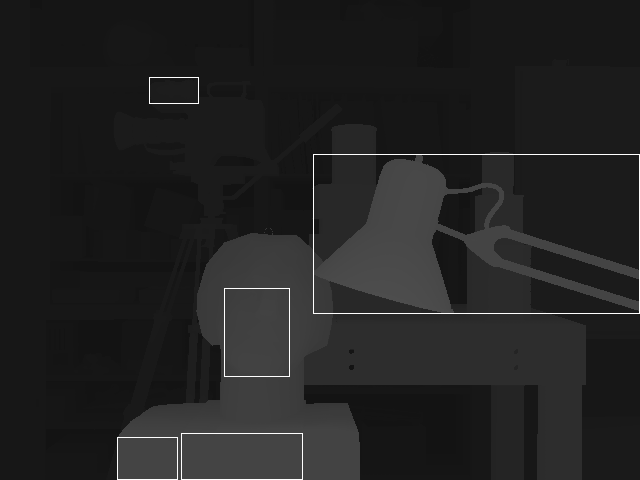}\strut
\end{minipage} & \begin{minipage}[t]{0.47\columnwidth}\raggedright\strut
Objects are recognized, the bust is recognized as 3 separate objects
with different disparity values since it actually has some volume, and
parts of it are deeper in the image. For now, the criteria of passing
this test is that we detected between 1 and 10 objects.\strut
\end{minipage}\tabularnewline
\begin{minipage}[t]{0.04\columnwidth}\raggedright\strut
4\strut
\end{minipage} & \begin{minipage}[t]{0.09\columnwidth}\raggedright\strut
A real calculated disparity image. It is noisier, so we set the minimum
object dimension to 40px for this image.\strut
\end{minipage} & \begin{minipage}[t]{0.29\columnwidth}\raggedright\strut
\includegraphics{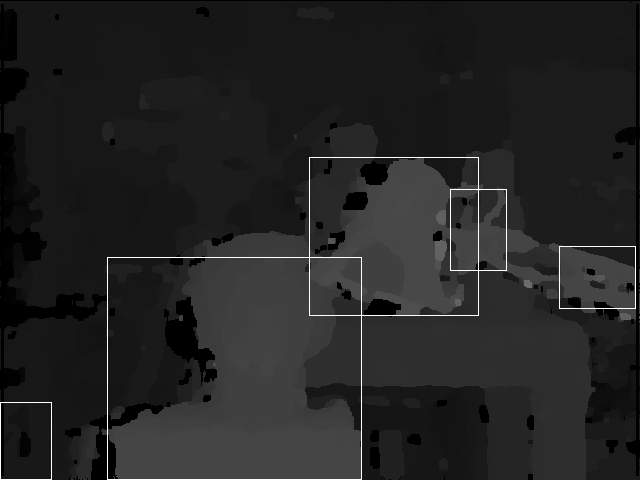}\strut
\end{minipage} & \begin{minipage}[t]{0.47\columnwidth}\raggedright\strut
The amount of detected blobs and their bounding boxes are adequate. For
now, the criteria of passing this test is that we detected between 1 and
100 objects.\strut
\end{minipage}\tabularnewline
\bottomrule
\end{longtable}

\section{Conclusion}\label{conclusion}

We have demonstrated an adaptive algorithm for performing Image
Segmentation on a non-binary image, such as a false-color (grayscale)
disparity image, using CCL. Our initial results show potential,
especially considering that they are portable to constrained systems
such as embedded ADAS. The current approach yields qualitatively better
results than OpenCV's \texttt{cv::SimpleBlobDetector}, which are
sensitive.

\subsection{Future Work}\label{future-work}

\begin{enumerate}
\def\labelenumi{\arabic{enumi}.}
\tightlist
\item
  Better control using the stop criteria. For instance, stopping when we
  have found \(N\) closest objects and none of them has ``grown'' in
  last \(M\) frames. This way we can balance between performance and the
  number of detected objects.
\item
  Implement
  \href{http://paper.ijcsns.org/07_book/200806/20080605.pdf}{Yapa \&
  Harada}'s proposal for at least one of the three mentioned CCL
  algorithms, and
  \href{http://imagelab.ing.unimore.it/yacclab}{benchmark} the quality
  and speed.
\item
  Consider alternatives to CCL for segmentation (Watershed, Region
  Growing, Clustering, etc.).
\item
  Improve the object-depth estimation algorithm beyond zeroth order
  (i.e.~taking the maximum disparity value in the bounding box).
\end{enumerate}

\section{References}\label{references}

\begin{enumerate}
\def\labelenumi{\arabic{enumi}.}
\tightlist
\item
  \href{https://www.sciencedirect.com/science/article/pii/S0031320317301693}{Lifeng
  He; Xiwei Ren; Qihang Gao; Xiao Zhao; Bin Yao; Yuyan Chao, ``The
  connected-component labeling problem: A review of state-of-the-art
  algorithms,'' Pattern Recognition, Elsevier BV, ISSN: 0031-3203, Vol:
  70, Page: 25-43 (2017), DOI: 10.1016/j.patcog.2017.04.018}
\item
  \href{http://www.cvlab.cs.tsukuba.ac.jp/dataset/tsukubastereo.php}{Martin
  Peris Martorell, Atsuto Maki, Sarah Martull, Yasuhiro Ohkawa, Kazuhiro
  Fukui, ``Towards a Simulation Driven Stereo Vision System''. ICPR2012,
  pp.1038-1042, 2012. Sarah Martull, Martin Peris Martorell, Kazuhiro
  Fukui, ``Realistic CG Stereo Image Dataset with Ground Truth Disparity
  Maps'', ICPR2012 workshop TrakMark2012, pp.40-42, 2012.}
\item
  \href{http://paper.ijcsns.org/07_book/200806/20080605.pdf}{Yapa, R.D.
  and Harada, K., 2008. ``Connected component labeling algorithms for
  gray-scale images and evaluation of performance using digital
  mammograms,'' International Journal of Computer Science and Network
  Security, 8(6), pp.33-41.}
\item
  \href{http://download.xuebalib.com/xuebalib.com.17233.pdf}{Suzuki, S.
  and Abe, K., ``Topological Structural Analysis of Digitized Binary
  Images by Border Following,'' CVGIP 30 1, pp 32-46 (1985)}
\item
  \href{https://ieeexplore.ieee.org/document/4310076}{Nobuyuki Otsu, ``A
  Threshold Selection Method from Gray-Level Histograms,''IEEE
  Transactions on Systems, Man, and Cybernetics, Volume 9, Issue 1,
  pp.~62 - 66 (Jan. 1979), DOI: 10.1109/TSMC.1979.4310076}
\item
  \href{http://imagelab.ing.unimore.it/yacclab}{Grana, Costantino;
  Bolelli, Federico; Baraldi, Lorenzo; Vezzani, Roberto ``YACCLAB - Yet
  Another Connected Components Labeling Benchmark'' Proceedings of the
  23rd International Conference on Pattern Recognition, Cancun, Mexico,
  pp.~3109 -3114 , Dec 4-8, 2016, DOI: 10.1109/ICPR.2016.7900112}
\end{enumerate}

\end{document}